\documentclass[journal]{IEEEtran}
\usepackage{cite}

\ifCLASSINFOpdf
\else
\fi
\usepackage{graphicx}
\usepackage{comment}
\usepackage{amsmath,amssymb} 
\usepackage{color}
\usepackage{bm}
\usepackage{multirow}

\usepackage{array}

\ifCLASSOPTIONcompsoc
 \usepackage[caption=false,font=normalsize,labelfont=sf,textfont=sf]{subfig}
\else
 \usepackage[caption=false,font=footnotesize]{subfig}
\fi

\newcommand{\red}[1]{\textcolor[rgb]{1,0,0}{#1}}
\newcommand{\blue}[1]{\textcolor[rgb]{0,0,1}{#1}}

\newcommand{\xy}[1]{\textcolor[rgb]{0,0,0}{#1}}

\begin{document}
%
\title{Weakly-Supervised Saliency Detection via Salient Object Subitizing}
%
%
%

\author{Xiaoyang Zheng*, \ \
        Xin Tan*, \ \
        Jie Zhou, \ \
        Lizhuang Ma$^\dagger$, \ \
        and \ \ Rynson W.H. Lau$^\dagger$
\thanks{* Equal Contribution.}
\thanks{$^\dagger$ Corresponding Author.}
\thanks{X. Zheng, X. Tan, J. Zhou and L. Ma are with the Department of Computer Science and Engineering, Shanghai Jiao Tong University, 200240, China. X. Tan and J. Zhou are also with the Department of Computer Science, City University of Hong Kong, HKSAR, China. E-mail:
zxyang.reg@sjtu.edu.cn,
tanxin2017@sjtu.edu.cn,
lord\_liang@sjtu.edu.cn,
ma-lz@cs.sjtu.edu.cn.}
\thanks{Rynson W.H. Lau is with the Department of Computer Science, City University of Hong Kong, HKSAR, China. E-mail: rynson.lau@cityu.edu.hk.}
\thanks{Manuscript received Sept. 20, 2020; revised Dec. 04, 2020.}}

%
%

\markboth{Journal of \LaTeX\ Class Files,~Vol.~14, No.~8, Sept.~2020}%
{Shell \MakeLowercase{\textit{et al.}}: Bare Demo of IEEEtran.cls for IEEE Journals}
%



\maketitle

\begin{abstract}
Salient object detection aims at detecting the most visually distinct objects and producing the corresponding masks. As the cost of pixel-level annotations is high, image tags are usually used as weak supervisions. However, an image tag can only be used to annotate one class of objects. In this paper, we introduce saliency subitizing as the weak supervision since it is class-agnostic. This allows the supervision to be aligned with the property of saliency detection, where the salient objects of an image could be from more than one class. To this end, we propose a model with two modules, Saliency Subitizing Module (SSM) and Saliency Updating Module (SUM). While SSM learns to generate the initial saliency masks using the subitizing information, without the need for any unsupervised methods or some random seeds, SUM helps iteratively refine the generated saliency masks. We conduct extensive experiments on five benchmark datasets. The experimental results show that our method outperforms other weakly-supervised methods and even performs comparable to some fully-supervised methods.
\end{abstract}


%
\IEEEpeerreviewmaketitle

\section{Introduction}
\label{sec:intro}
The salient object detection task aims at accurately recognizing the most distinct objects in a given image that would attract human attention. This task has received a lot of research interests in recent years, as it plays an important role in many other computer vision tasks, such as visual tracking \cite{qin2017real}, image editing/manipulation \cite{goferman2011context,cheng2010repfinder} and image retrieval \cite{he2012mobile}.
Recently, deep convolutional neural networks have achieved significant progress in saliency detection {\cite{he2015supercnn,liu2016dhsnet,zhuge2019deep,zhu2019aggregating,hu2020sac}.}
However, most of these recent methods are primarily CNN-based, which rely on a large amount of pixel-wised annotations for training. For such an image segmentation task, it is both arduous and inefficient to collect a large amount of pixel-wised saliency masks. For example, it usually takes several minutes for an experienced worker to annotate one single image. This drawback confines the amount of available training samples. In this paper, we focus on the salient object detection task with a weakly-supervised setting.

\begin{figure}
    \centering
    \includegraphics[width=0.5\textwidth]{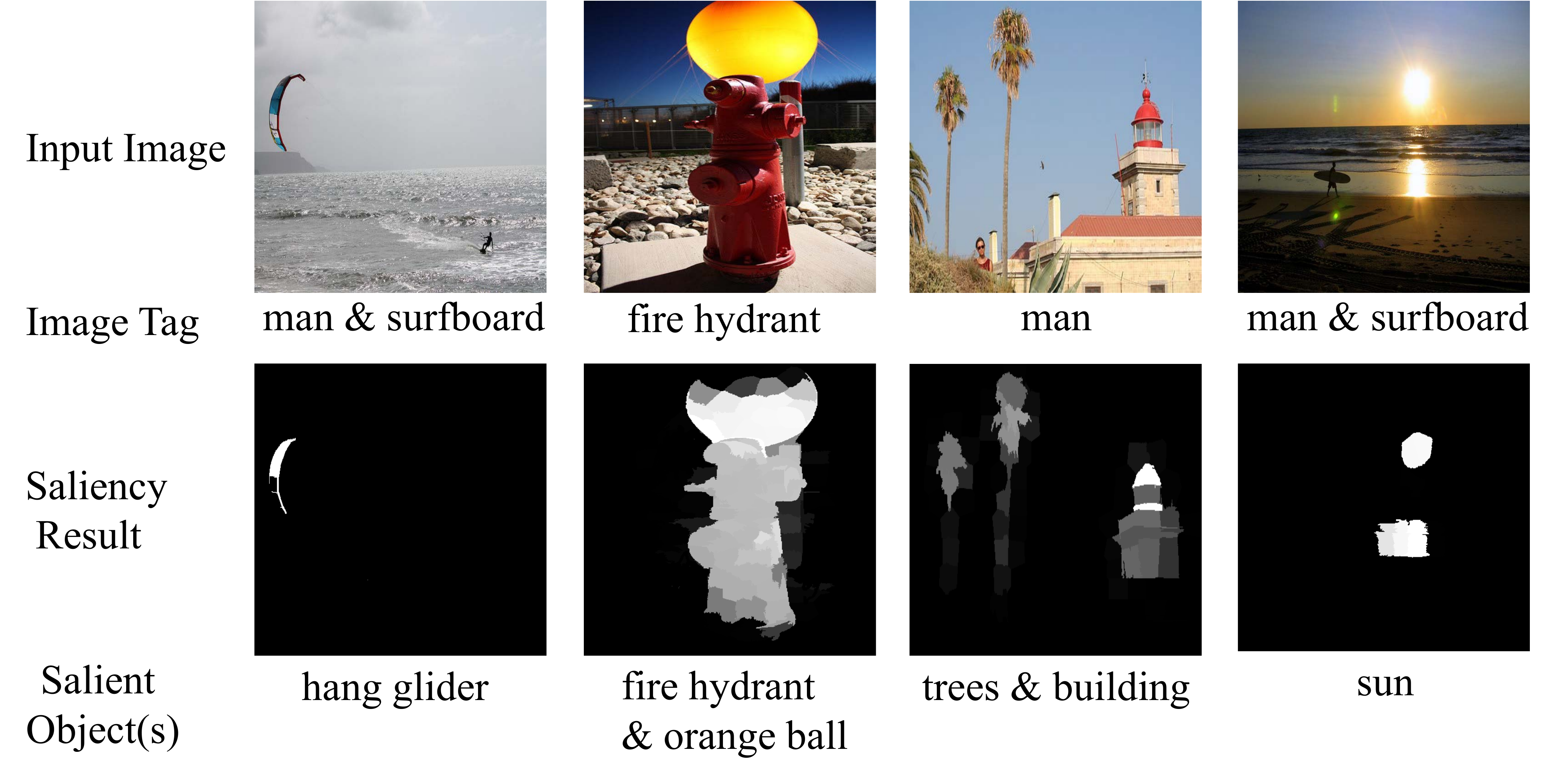}
    \caption{Several inconsistent cases between the given image labels and the actual salient objects. These images and tags are chosen from the Pascal VOC \cite{everingham2010pascal} and Microsoft COCO \cite{lin2014microsoft} datasets. The captions under images are the given labels. The masks are generated with our methods, which show the actually salient objects.}
    \label{fig:diss}
\end{figure}

Some methods \cite{li2018weakly,wang2017learning,zeng2019multi} tried to address salient object detection with image-tag supervisions. 
Li et~al. \cite{li2018weakly} utilized Class Activation Maps as coarse saliency maps. Together with results from unsupervised methods, those masks are used to supervise the segmentation stream. Wang et~al \cite{wang2017learning} proposed a two-stage method, which assigns category tags to object regions and fine-tunes the network with the predicted saliency maps with the ground truth. Zeng et~al. \cite{zeng2019multi} proposed a unified framework to conduct saliency detection with diverse weak supervisions, including image tags, captions and pseudo labels. 
They achieved good performances with image labels from the Microsoft COCO \cite{lin2014microsoft} or Pascal VOC \cite{everingham2010pascal} datasets. However, their results are established on a critical assumption that \textit{the labelled object is the most visually distinct one.} From those datasets with image tags, We observe that \textit{this assumption is not always reliable}. As shown in Figure~\ref{fig:diss}, the actual salient objects are inconsistent with the image labels. For example, the image in the second column is labelled as ``fire hydrant'', 
it is obvious that the orange ``ball'' should also be a salient object. 
In addition, even trained on datasets with multi-class labels, these methods essentially detect object within the categories, but not salient objects.
Hence, image category labels do not guarantee the property of saliency. 

\textbf{Motivation.} Subitizing is the rapid enumeration of a small number of items in a given image, regardless of their semantic category. According to \cite{zhang2017salient_jour}, subitizing of up to four targets is highly accurate, quick and confident. 
In addition, since the subitizing information may contain objects from different categories, it is class-agnostic.
Inspired by the above advantages, we propose to address the saliency detection problem using only the object subitizing information as a weak supervision.

Although there exist works, e.g., \cite{he2017delving}, that use subitizing as an auxiliary supervision, we propose to apply subitizing as the weak supervision in this work.
To the best of our knowledge, we are the first to adopt subitizing as the only supervision in saliency detection task. However, the subitizing information does not indicate the position and appearance of salient objects. Therefore, we propose the Saliency Subitizing Module (SSM) to produce saliency masks. Recent works \cite{zhou2016learning,selvaraju2017grad} have proven that, even trained with classification tasks, 
CNNs implicitly reveal the attention regions in the given images.
Trained on subitizing information, the SSM relies on the distinct regions to conduct classification. By extracting those regions, we can explicitly obtain the locations of the salient objects.

However, as pointed out in \cite{woo2018cbam}, in a well trained classification network, the discriminative power of each object part is different from each other, and thus lead to incomplete segmentation. In the finetune stage, we need to further enlarge the prominent regions extracted from the network. Kolesnikov et~al. \cite{kolesnikov2016seed} trained their network with pseudo labels for multiple iterations and obtain integrated results, while the multi-stage training is complicated.
In order to address this issue, we design the Saliency Updating Module (SUM) for refining the saliency masks produced by SSM.
In each iteration, the generated saliency maps, combined with original images, are used to generate masked images. 
With those masked images as input to the next iteration, the network learns to recognize those related but less salient regions. 
During the inference phase, given an image, our model will produce the saliency maps without any iterations, and there will be no need to provide the subitizing information. 
Our extensive evaluations demonstrate the superiority of the proposed methods over the state-of-the-art weakly-supervised methods.

In summary, this paper has the following contributions:
\begin{itemize}
    \item We propose to use subitizing as a weak supervision in the saliency detection task, which has the advantage of being class-agnostic.
    \item 
    We propose an end-to-end multi-task saliency detection network. By introducing subitizing information, our network first generates rough saliency masks with the Saliency Subitizing Module (SSM), and then iteratively refines the saliency masks with the Saliency Updating Module (SUM).
    \item 
    Our extensive experiments show that the proposed method achieves superior performance on five popular salient datasets, compared with other weak-supervised methods. It even performs comparable to some of the fully-supervised methods.
\end{itemize}


\section{Related Work}
\label{sec:related}
Recently, the progress on salient object detection is substantial, benefiting by the development of deep neural networks. He et~al. \cite{he2015supercnn} proposed a convolution neural network based on super-pixel for saliency detection. Li et~al. \cite{li2015visual} utilized multi-scale features extracted from a deep CNN. Zhao et~al. \cite{zhao2015saliency} proposed a multi-context deep learning framework for detecting salient objects with two different CNNs used to learn global and local context information. \xy{Yuan et~al. \cite{yuan2016dense} proposed a saliency detection framework, which extracted the correlations among macro object contours, RGB features and low-level image features.} Wang et~al. \cite{wang2019salient} proposed a pyramid attention structure to enlarge the receptive field. Hou et~al. \cite{HouCHBTT19} introduced short connections to an edge detector. \xy{Zhu et~al. \cite{zhu2019aggregating} proposed the attentional DenseASPP to exploit local and global contexts at multiple dilated convolutional layers. Hu et~al. \cite{hu2020sac} proposed a spatial attenuation context network, which recurrently translated and aggregated the context features in different layers.
Tu et~al. \cite{tu2020edge} presented an edge-guided block to embed boundary information into saliency maps.
} Zhou et~al. \cite{zhou2020multi} proposed a multi-type self-attention network to learn more semantic details from degraded images. However, these methods rely heavily on pixel-wised supervisions. Due to the scarcity of pixel-wised data, we focus on the weakly-supervised saliency detection task.

\textbf{Weakly-Supervised Saliency detection.} 
There are many works using weak supervisions for the saliency detection task.
For example, Li et~al. \cite{li2018weakly} used the image-level labels to train the classification network and applied the coarse activation maps as saliency maps. Wang et~al. \cite{wang2017learning} proposed a two-stage weakly-supervised method by designing a Foreground Inference Network (FIN) to predict foreground regions and a Global Smooth Pooling (GSP) to aggregate responses of predicted objects. 
Zeng et~al. \cite{zeng2019multi} proposed a unified network, which is trained on multi-source weak supervisions, including image tags, captions and pseudo labels. They designed an attention transfer loss to transmit signals between sub-networks with different supervisions. However, as discussed in Section~\ref{sec:intro}, the image-level supervisions are not always reliable. In addition, captions were used as a weak supervision in \cite{zeng2019multi}, combined with other supervisions. 
Different from those methods above, we propose to use subitizing information as the weak supervision in the saliency detection task.

\textbf{Salient object subitizing.} Zhang et~al. \cite{zhang2015salient} proposed a salient object subitizing dataset SOS. They firstly studied the problem of salient object subitizing and revealed the relations between subitizing and saliency detection. Lu et~al. \cite{lu2018class} formulated the subitizing task as a matching problem and exploited the self-similarity within the same class. He et~al. \cite{he2017delving} trained a subitizing network to provide additional knowledge to the pixel-level segmentation stream. Recently, Amirul et~al. \cite{amirul2018revisiting} proposed a salient object subitizing network and recognized the variability of subitizing. They also provided outputs as a distribution that reflects this variability. In this paper, our approach is motivated by these methods but we use subitizing as the weak supervision.  


\textbf{Map refinement.}
Li et~al. \cite{li2018weakly} adopted saliency maps generated by some unsupervised methods as the initial seeds. With a graphical model, these saliency maps are used as pixel-level annotations and refined in an iterative way. However, in our proposed method, we do not utilize any unsupervised methods or initial seeds. The saliency maps are refined iteratively from the activation maps of the subitizing network. 
\xy{Li et~al. \cite{li2018tell} adopted a soft mask loss to an auxiliary attention stream. However, the input of \cite{li2018tell} is updated only once while the inputs to our network are iteratively updated.}
In addition, there are some existing post-processing techniques used to refine the saliency masks. In \cite{wang2017learning,li2018weakly}, \xy{the authors} utilized an iterative conditional random field (CRF) to enforce spatial label consistency. Zheng et~al. \cite{zheng2015conditional} further proposed to conduct approximate inference with pair-wised Gaussian potential in CRF as a recurrent neural network. Chen et~al. \cite{chen2020learning} employed the relations of the deep features to promote saliency detectors in a self-supervised way. In order to achieve better results, we adopt a refinement process, which maintains the internal structure of original images and enforces smoothness into the final saliency maps.

\begin{figure*}[htbp]
    \centering
    \includegraphics[width=0.9\textwidth]{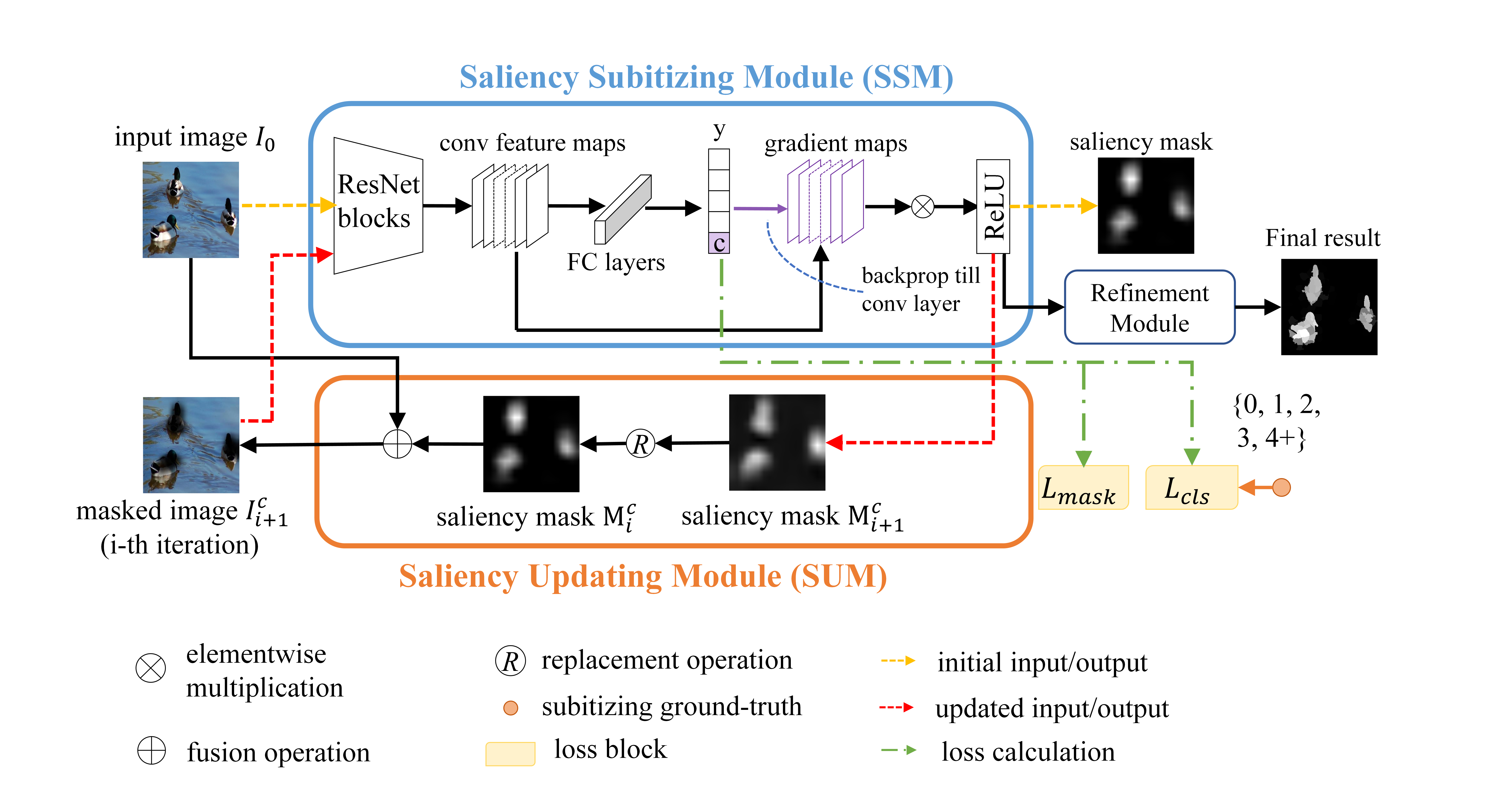}
    \caption{\xy{The pipeline of the proposed network, with the Saliency Subitizing Module (SSM), the Saliency Updating Module (SUM) and the refinement process.}}
    \label{fig:network}
\end{figure*}

\section{Salient Object Detection Method}
\label{sec:method}


We propose a multi-task convolutional neural network, which consists of two main modules: saliency subitizing module (SSM) and saliency updating module (SUM). SSM helps learn counting of salient objects and extract coarse saliency maps with the precise locations of the target objects. SUM helps update saliency masks produced by SSM, and extend the activation regions. Finally, we apply a refinement process to refine the object boundaries. The pipeline of our method is presented in Figure~\ref{fig:network}.


\subsection{Saliency Subitizing Module}
The subitizing information indicates the number of salient objects in a given image. It does not provide the location or the appearance information of the salient objects explicitly. However, when we train our network with the subitizing supervisons, the network will learn to focus on regions related to the salient objects. Hence, we design the Saliency Subitizing Module (SSM) to extract the attention regions as coarse saliency masks. We regard the saliency object subitizing task as a classification task. Training images are divided into 5 categories based on the number of salient objects: $0$, $1$, $2$, $3$ and $4+$. We use ResNet-50 \cite{he2016deep} as the backbone network, pretrained on the ImageNet \cite{deng2009imagenet} dataset. The original 1000-way fully-connected layer are replaced by a 5-way fully-connected layer. We use cross-entropy loss as the classification loss $L_{cls}$. In order to obtain denser saliency maps, the stride of the last two down-sampling layers is set as 1 in our backbone network, which produces feature maps with $1/8$ of the original resolution before the classification layer. In order to enhance the representation power of the proposed network, we also apply two attention modules: channel attention module and spatial attention module, which tells the network ``where'' and ``what'' to focus, respectively. Both of them are placed in a sequential way between the ResNet blocks.

We apply the technique of Grad-CAM \cite{selvaraju2017grad} to extract activation regions as the initial saliency maps, which contains the gradient information flowing into the last convolutional layers during the backward phase. The gradient information represents the importance of each neuron during the inference of the network. We assume that the produced features from the last convolutional layer have a channel size of $K$. For a given image, let $f_k$ be the activation of unit $k$, where $k\in [1, K]$. For each class $c$, the gradients of the score $y^c$ with respect to the activation map $f_k$ are averaged to obtain the neuron importance weight $\alpha_k^c$ of class $c$:
\begin{equation}
    \alpha_k^c=\frac{1}{N}\sum_i^m\sum_j^h\frac{\partial y^c}{\partial f_{ij}^k},
\end{equation}
where $i$ and $j$ represent the coordinates in the feature map and $N=m\times h$. With the neuron importance weight $\alpha_k^c$, we can compute the activation map $M^c$:
\begin{equation}
    M^c=\text{ReLU}(\sum_k \alpha_k^c f^k).
\end{equation}
Note that the \text{ReLU} function filters the negative gradient values, since only the positive ones contribute to the class decision, while the negative values contribute to other categories. The size of the saliency map is the same as the size of the last convolutional feature maps ($1/8$ of the original resolution). Since the saliency maps $M^c$ are obtained within each inference, they become trainable during the training stage.

\subsection{Saliency Updating Module}



In the Saliency Subitizing Module, our proposed network learns to detect the regions that contribute to the counting of salient objects. Due to this attribute, with only the SSM module, we can obtain saliency masks with accurate locations of the target objects.
However, the quality of the saliency masks may not be very high. In order to address this issue, 
we design a Saliency Updating Module (SUM) to fine-tune the obtained masks, with the aim to refine the activation regions. Li et~al. \cite{li2018weakly} updated saliency mask with an additional CRF module. In contrast, our proposed model refines the saliency masks in an end-to-end way.


As shown in Figure~\ref{fig:network}, we fuse the origin images and the saliency maps to obtain masked images as new inputs to the next iteration. Visually, the current prominent area is eliminated from the original samples. We define $I_0$ as the original images. $I^c_i$ denotes the input images at the $i$-th iteration ($i>=1$). $M^c_i$ denotes the saliency maps of class $c$ at the $i$-th iteration. 

The fusion operation is formulated as:
\begin{equation}
    I^c_i = I_0-(\text{Sigmoid}(\omega\cdot(M^c_{i-1}-\bm{\sigma}))\odot I_0),
    \label{eq:mask}
\end{equation}
where $\odot$ stands for element-wise multiplication,
$\bm{\sigma}$ is a threshold matrix with all elements equal to $\sigma$. With the scale parameter $\omega$, the mask term gets closer to 1 when $M^c_{i-1}>\sigma$, and gets closer to 0 otherwise. As presented in Eq.~\ref{eq:mask}, 
we enforce $I_i^c$ to contain as few features from the target class $c$ as possible.

Trained on samples without features from the current prominent area, the network learns to recognize those related but less salient regions. In other words, regions beyond the high-responding area should also include features that trigger the network to recognize the sample as class $c$. \xy{Similar to \cite{li2018tell},} 
we introduce an mask mining loss $L_{mask}$ to extract larger activation area. This loss \xy{penalizes the prediction error for class} $c$, as shown below,
\begin{equation}
    L_{mask}=\frac{1}{n}\sum_c y^c(I^{c}),
    \label{eq:mask loss}
\end{equation}
where $n$ is the \xy{dataset size} and $y^c(I^{c})$ represents the prediction score of masked images $I^{c}$ for class $c$. With the loss perspective, the prediction scores of the right label for the masked images should be lower than those for the original images. 
During the training phase, the total loss $L$ is the combination of the classification loss $L_{cls}$ and the mask mining loss $L_{mask}$.
\begin{equation}
    L=L_{cls}+\alpha L_{mask},
    \label{eq:loss}
\end{equation}
where $\alpha$ is the weighting parameter. We set $\alpha=1$ in all our experiments. With this loss, the network learns to extract those less salient but related parts of the target objects, while maintains the ability of recognizing subitizing information.
\xy{In \cite{li2018tell}, the extracted regions for masking the input are updated only once. These regions extracted from just a single step are usually small and sparse, since CNNs tend to capture the most discriminative regions and neglect the others in the image.
In contrast, our method updates the extracted regions through multiple iterations. In this way, the extracted regions in our method are more integrated.}

\xy{\textbf{Discussion.} Although Wei et al. \cite{wei2017object} also adopted an iterative strategy,
our method is different from \cite{wei2017object} in two main aspects. First, during the generation of training images for the next iteration, \cite{wei2017object} simply replaced the internal pixels by the mean pixel values of all the training images. Instead, we use a threshold $\sigma$ to determine the salient regions and a weighting parameter $\omega$ to adjust the removing rate of features (as presented in Eq. \ref{eq:mask}), 
so that the correlations of the extracted regions and the backgrounds at different iterations would be smoothly changed.
Second, \cite{wei2017object} took the mined object region as the pixel-level label for segmentation training.
Instead, our method is only trained on the given dataset with subitizing labels, avoiding training on unreliable pseudo labels.}

\subsection{Refinement Process}

To refine the object boundaries in the saliency maps, we take a graph-based optimization method. Inspired by \cite{li2016deepsaliency}, we adopt super-pixels produced with SLIC \cite{achanta2009frequency} as the basic representation units. Those super-pixels are organized as an adjacency graph to model the structure in both spatial and feature dimensions. A given image is segmented into a set of super-pixels $\{x_i\}_{i=1}^N$, where $N=200$.
The super-pixel graph $A=(a_{i,j})_{N\times N}$ is defined as follows:
\begin{equation}
    a_{i,j}=\left\{
    \begin{array}{ll}
        K(x_i,x_j), & \text{if $x_i$ and $x_j$ are spatially adjacent;} \\
        0, & \text{otherwise,}
    \end{array}
    \right .
\end{equation}
where $K(x_i,x_j)=\text{exp}(-\frac{1}{2}\|x_i-x_j\|^2_2)$ evaluates the feature similarity. 
Assume that there exist $l$ super-pixels with initial scores. Our task is to learn a non-linear regression function $g(x)=\sum_{j=1}^N\alpha_j K(x_i,x_j)$ for each super-pixel $x$. The framework is shown as:
\begin{equation}
\min_g \frac{1}{l}\sum_{i=1}^l (y_i-g(x_i))^2+\theta_1 \|g\|_K^2+\frac{\theta_2}{N^2}g^T D^{-\frac{1}{2}}LD^{-\frac{1}{2}}g,
\label{eq:eq8}
\end{equation}
where $y_i$ is the weight of the i-th unit in the super-pixel graph, and $g=(g(x_1), ..., g(x_N))^T$. $\|g\|_K$ denotes the norm of $g$ induced by the function $K$; $D$ is the diagonal matrix containing the degree value in the adjacency graph; $L$ denotes the graph Laplacian matrix, defined as $L=D-A$; $\theta_1$ and $\theta_2$ are two weights, set as $1$ and $1e-6$, respectively.

\begin{figure}
    \centering
    \includegraphics[width=0.5\textwidth]{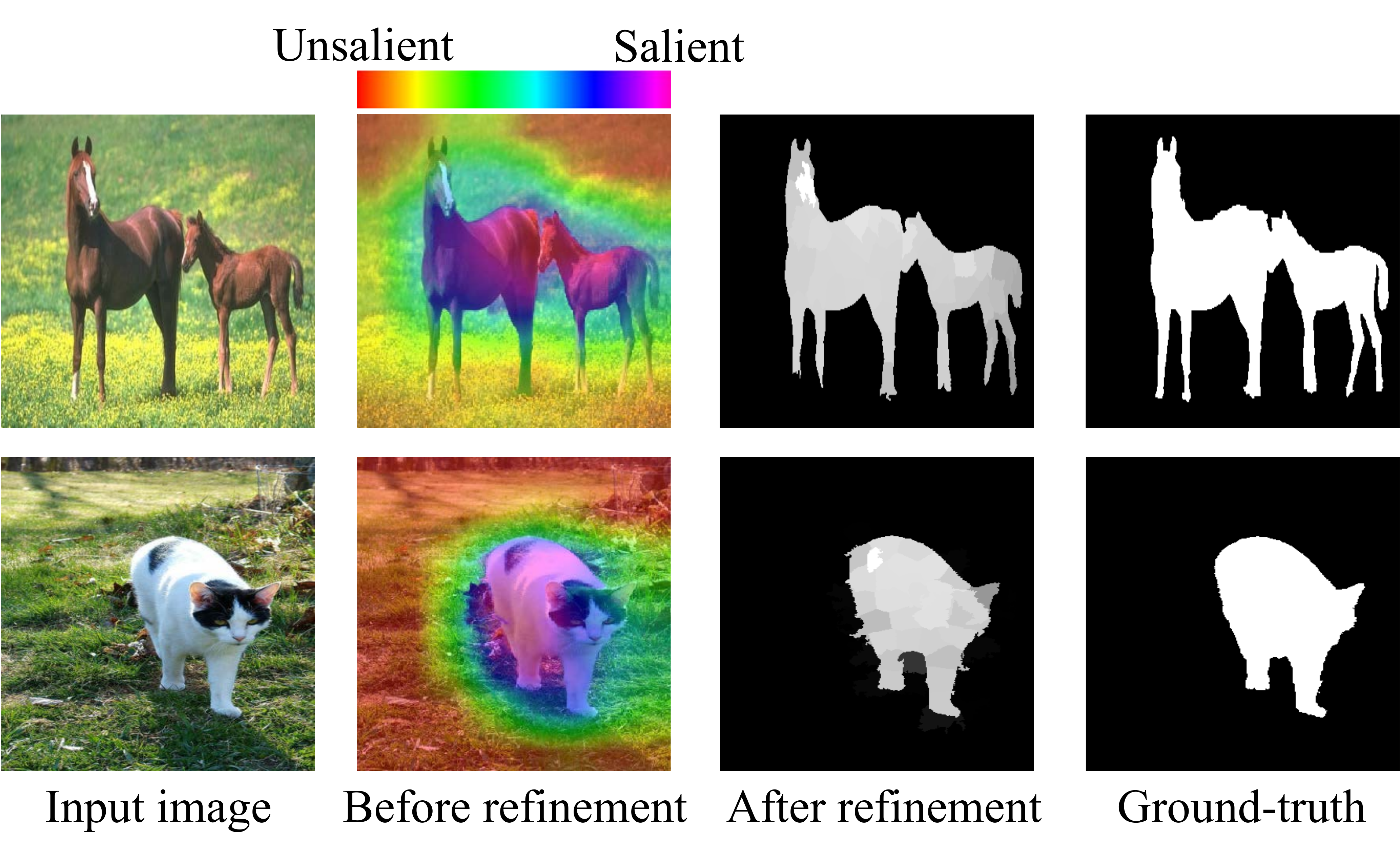}
    \caption{Comparison between coarse saliency maps and refined saliency maps. The color code above the second column indicates the degree of saliency.}
    \label{fig:refine}
\end{figure}

\xy{In Eq.~\ref{eq:eq8}, the first term is the trivial square loss, and the second term aims at normalizing the desired regression function. 
However, unlike \cite{li2016deepsaliency}, we introduce matrix $D$ in the third term to enforce constraints between units of the super-pixel graph, and normalize the optimized results. Since we introduce constraints between different graph units, our method can help strengthen the connections and smoothness among different graph units.} The optimization objective function is transformed into a matrix form as:
\begin{equation}
\min_{\alpha}\frac{1}{l}\sum_{i=1}^l\|y-JK\alpha\|_2^2+\theta_1\alpha^TK\alpha+\frac{\theta_2}{N^2}\alpha^T KD^{-\frac{1}{2}}LD^{-\frac{1}{2}}K\alpha,
\end{equation}
where $J$ is a diagonal matrix with the first $l$ elements set to $1$, while the other elements are set to $0$; 
$\alpha=(\alpha_1,\ldots,\alpha_N)^T$; $K$ is the kernel gram matrix. The solution to the above optimization problem is formulated as:
\begin{equation}
\alpha^*=(JK+\theta_1 lI+\frac{\theta_2l}{N^2}D^{-\frac{1}{2}}LD^{-\frac{1}{2}}K)^{-1}y,
\end{equation}
where $I$ is an identity matrix. 
With the optimized $\alpha^*$, we can calculate the saliency score $g(x)$ for each super-pixel. As presented in Figure~\ref{fig:refine}, the refinement process optimizes the boundaries of the salient maps.

\section{Experiment Results}
\label{sec:experiment}

\subsection{Implementation Details}
In this paper, we utilize ResNet-101 \cite{he2016deep} as the backbone and modify it to meet our requirement. Those unmodified layers are initialized with weights pretrained on ImageNet \cite{deng2009imagenet}, while the rest are randomly initialized. Our method is implemented based on the PyTorch framework. We use Stochastic Gradient Descent for parameter updating. The learning rate is initially set up as 1e-3, and will go down as the training progresses. The weight decay and the momentum is set as 5e-4 and 0.9, respectively. 

It has been widely proved that inputs with various scales helps the accurate localization of target objects. 
Hence, the input scales are set as $\{0.5, 0.75, 1.0\}$ of the original size. Saliency maps from three replicate networks with shared weights are summed up and regularized as $[0, 1]$. The proposed method is trained and evaluated on a PC with i9 3.3GHz CPU, an Nvidia 1080Ti GPU and 64GB RAM. Given an image of 400$\times$400, the network takes about 0.05s to produce a single saliency map and the refinement procedure takes 0.03s per image. We apply random horizontal flipping, color scale and random rotations ($\pm 30^\circ$) to augment the training datasets.



\subsection{Datasets and Evaluation Metrics}

\xy{The ESOS dataset \cite{zhang2017salient_jour} is a saliency detection dataset, annotated with subitizing labels. It contains $17,000$ images, which are selected from four datasets: MS COCO \cite{lin2014microsoft}, Pascal VOC 2007 \cite{everingham2010pascal}, ImageNet \cite{deng2009imagenet} and SUN \cite{xiao2010sun}.}
Each image in the dataset is \xy{re-labeled} by \cite{zhang2017salient_jour} with $0$, $1$, $2$, $3$ and $4+$ salient objects (5 classes). \xy{We randomly choose $80\%$ of the whole dataset (around $13,000$ images) as the training set and use the rest $20\%$ as the validation set for model selection. All images are scaled to $256\times256$ during training.} 
To compare with other saliency detection methods, we evaluate our proposed method on five benchmarks: Pascal-S \cite{li2014secrets}, ECSSD \cite{shi2015hierarchical}, HKU-IS \cite{li2016deep}, DUT-OMRON \cite{yang2013saliency} and MSRA-B \cite{jiang2013salient}. These five datasets are all commonly used in the saliency detection task. All of them provide images and pixel-level masks.

In this paper, we adopt four metrics to measure the performance: precision-recall (PR) curve, $F_\beta$, S-measure \cite{fan2017structure} and mean absolute error (MAE). The continuous saliency maps are binarized with different threshold values. The PR curve is computed by comparing a series of binary masks with the ground truth masks. The second metric is defined as:
\begin{equation}
    F_\beta=\frac{(1+\beta^2)\cdot\textit{precision}\cdot\textit{recall}}{\beta^2\cdot\textit{precision}+\textit{recall}},
\end{equation}
where $\beta^2$ is set as $0.3$ to balance between precision and recall \cite{achanta2009frequency}. The maximum F-measure is selected among all precision-recall pairs. The structural measure, or S-measure \cite{fan2017structure}, is used to evaluate the structural similarity of non-binary saliency maps. It is defined as:
\begin{equation}
    S = \frac{S_o + S_r}{2},
\end{equation}
$S_o$ is used to assess the object structure similarity against the ground truth, while $S_r$ measures the global similarity of the target objects.
Please refer to the original paper \cite{fan2017structure} for the definition of these two terms.
The metric of MAE measures the average pixel-wised absolute difference between the binary ground-truth and the saliency maps. It is calculated as:
\begin{equation}
    {MAE}(S, GT)=\frac{1}{W\times H}\sum^W_{x=1}\sum^H_{y=1}\|S(x,y)-GT(x,y)\|,
\end{equation}
where $S$ and $GT$ are generated saliency maps and the ground truth of size $W\times H$.

\begin{figure*}[htbp]
	\centering
	\includegraphics[width=\textwidth]{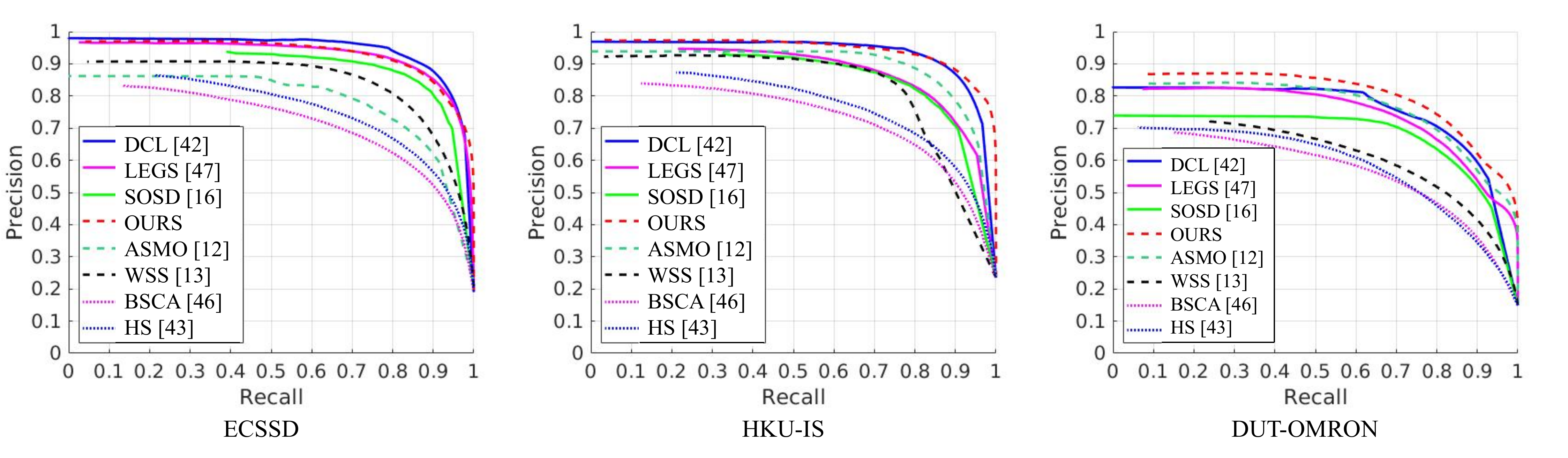}
	\caption{Precision-recall curves of our proposed method and other methods on three benchmark datasets. Our proposed method consistently outperforms unsupervised and weakly-supervised methods. It is comparable to some fully-supervised methods.}
	\label{fig:pr}
\end{figure*}

\begin{table*}[htbp]
    \tiny
    \centering
    \caption{Quantitative results on $F_{\beta}$ and MAE.
    The \red{red} ones refer to the best results and the \blue{blue} ones refer to the second best results.}\label{tab:comparison}
    \begin{tabular}{|c|c|c|c|c|c|c|c|c|c|c|}
        \hline
        \multirow{2}{*}{Dataset} & \multirow{2}{*}{Metric} & \multicolumn{2}{|c|}{Unsupervised} & \multicolumn{3}{|c|}{Weakly-supervised} & \multicolumn{4}{|c|}{Fully-supervised}  \\
        \cline{3-11}
        ~ & ~ &
        BSCA\cite{qin2015saliency} & HS\cite{yang2013saliency} &
        ASMO\cite{li2018weakly} & WSS\cite{wang2017learning} & \textbf{Ours} & SOSD\cite{he2017delving} & LEGS\cite{wang2015deep} & DCL\cite{li2016deep} &
        MSWS\cite{zeng2019multi} \\
        \hline
        \multirow{2}{*}{ECSSD} & $F_{\beta}\uparrow$ & 0.705 & 0.727 & 0.837 & 0.823 & \textbf{\blue{0.858}} & 0.832 & 0.827 & 0.859 & \textbf{\red{0.878}}\\
        \cline{2-11}
    ~ & MAE$\downarrow$ & 0.183 & 0.228 & 0.110 & 0.120 & 0.108 & \textbf{\blue{0.105}} & 0.118 & 0.106 & \textbf{\red{0.096}} \\\hline
        \multirow{2}{*}{MSRA-B} & $F_{\beta}\uparrow$ & 0.830 & 0.813 & 0.881 & 0.845 & \textbf{\blue{0.897}} & 0.875 & 0.870 & \textbf{\red{0.905}} & 0.890 \\
        \cline{2-11}
        ~ & MAE$\downarrow$ & 0.131 & 0.161 & 0.095 & 0.112 & 0.082 & 0.104 & \textbf{\blue{0.081}} & 0.072 & \textbf{\red{0.071}} \\\hline
        \multirow{2}{*}{DUT-OMRON} & $F_{\beta}\uparrow$  &  0.500 & 0.616 & 0.722 & 0.657 & \textbf{\red{0.778}} & 0.665 & 0.669 & \textbf{\blue{0.733}} & 0.718 \\
        \cline{2-11}
        ~ & MAE$\downarrow$  & 0.196 & 0.227 & 0.110 & 0.150 & \textbf{\red{0.083}} & 0.198 & 0.133 & \textbf{\blue{0.094}} & 0.114 \\\hline
        \multirow{2}{*}{Pascal-S} & $F_{\beta}\uparrow$ &  0.597 & 0.641 & 0.752 & 0.720 & \textbf{\blue{0.803}} & 0.794 & 0.752 & \textbf{\red{0.815}} & 0.790\\
        \cline{2-11}
        ~ & MAE$\downarrow$  & 0.225 & 0.264 & 0.152 & 0.145 & \textbf{\blue{0.131}} & 0.114 & 0.157 & \textbf{\red{0.113}} & 0.134 \\\hline
        \multirow{2}{*}{HKU-IS} & $F_{\beta}\uparrow$  & 0.654 & 0.710 & 0.846 & 0.821 & \textbf{\blue{0.882}} & 0.860 & 0.770 & \textbf{\red{0.892}} & / \\
        \cline{2-11}
        ~ & MAE$\downarrow$  & 0.174 & 0.213 & 0.086 & 0.093 & \textbf{\blue{0.080}} & 0.129 & 0.118 & \textbf{\red{0.074}} & / \\\hline
    \end{tabular}

\end{table*}

\subsection{Visualized Results}
\begin{figure*}
    \centering
    \includegraphics[width=0.85\textwidth]{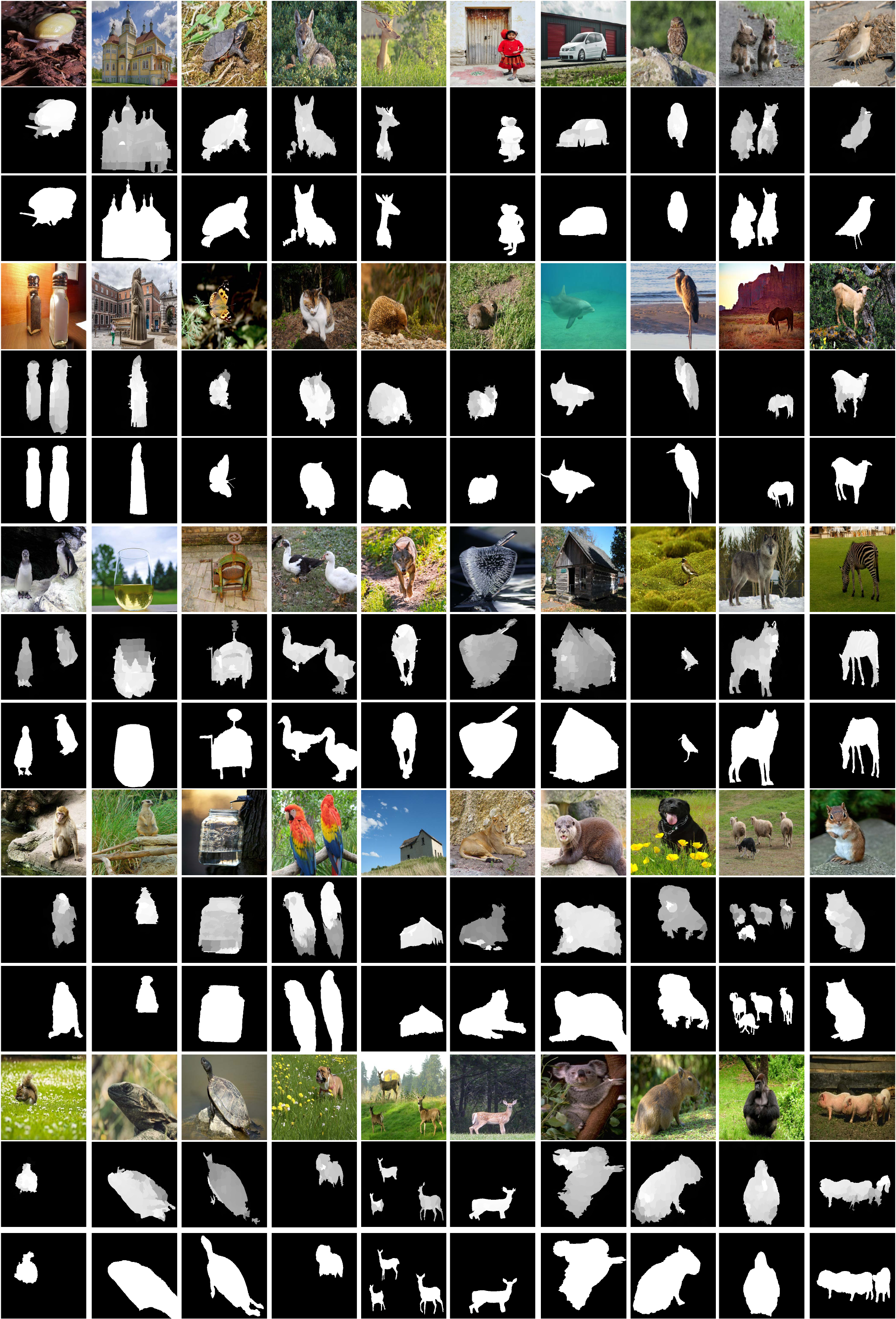}
    \caption{Visualized results of our proposed method. From top to bottom, there are five groups, organized as: images, our results and the ground-truth.}
    \label{fig:vis}
\end{figure*}

We present some visualized results of our proposed method in Figure \ref{fig:vis}, which is close to the ground-truth.

\subsection{Comparison with Other Methods}

We conduct the saliency detection task in a weakly-supervised way. There exist several other weakly-supervised methods. The difference on the settings is presented in Table~\ref{tab:setting}. Our method requires less extra information than those existing weakly-supervised methods. In addition, we apply the refinement process to optimize the saliency maps, instead of the commonly used CRF. 

\begin{table}
    \small
    \centering
    \caption{Comparison of our setting with other methods. \textit{Seed} means using unsupervised saliency maps as the initial seeds. \textit{Pixel} means applying pixel-wised supervision. \textit{CRF} means using conditional random fields as the post-processing step. \textit{Pseudo} means adopting pseudo labels.}
    \label{tab:setting}
    \begin{tabular}{|c|c|c|c|c|}
    \hline
        Setting & ASMO \cite{li2018weakly} & WSS \cite{wang2017learning} & SOSD \cite{he2017delving} & Ours \\\hline
        Label & image tag & image tag & subitizing+pixel & subitizing \\\hline
        Seed & \checkmark & $\times$ & $\times$ & $\times$ \\\hline
        Pixel & $\times$ & $\times$ & \checkmark & $\times$ \\\hline
        Pseudo & \checkmark & \checkmark & $\times$ & $\times$ \\\hline
        CRF & \checkmark & \checkmark & \checkmark & $\times$ \\
    \hline
    \end{tabular}
\end{table}


We compare our results with eight state-of-the-art methods, including two unsupervised ones: BSCA\cite{qin2015saliency} and HS\cite{yang2013saliency}; two weakly-supervised ones using image-label supervisions: ASMO\cite{li2018weakly} and WSS\cite{wang2017learning}; four fully-supervised ones: SOSD\cite{he2017delving}, LEGS\cite{wang2017learning}, DCL\cite{li2016deep} and MSWS\cite{zeng2019multi}.

As shown in Table~\ref{tab:comparison}, our proposed method outperforms existing unsupervised methods with a considerable margin. Compared to weakly-supervised methods with image-label supervisions, our method achieves better performance on all benchmarks. It proves that the subitizing supervision helps boost the saliency detection task. In addition, our method compares favorably against some fully-supervised counterparts. Note that on the DUT-OMRON dataset, our method obtains more precise results than the fully-supervised methods. Since the masks of the DUT-OMRON dataset are complex in appearance, sometimes with holes, it reveals that our method is capable of handling difficult situations. Compared to SOSD \cite{he2017delving}, which utilized additional subitizing information, our method extracts more valid information from the subitizing supervision. Moreover, our method achieve comparable results with MSWS \cite{zeng2019multi}, which applied multi-source weak supervisions, including subitizing, image labels and captioning.

\begin{table}[htbp]
	\centering
    \caption{Comparison of our results with two weakly-supervised methods (ASMO \cite{li2018weakly} and WSS \cite{wang2017learning} using image-tag supervision) and a fully-supervised method SOSD \cite{he2017delving} additionally using subitizing information in terms of S-measure (larger is better).}
    \label{tab:s}
	\begin{tabular}{|c|c|c|c|c|}
	\hline
		Methods & ASMO\cite{li2018weakly} & WSS\cite{wang2017learning} & SOSD\cite{he2017delving} & Ours \\\hline
		ECSSD & 0.827 & 0.829 & 0.837 & \textbf{0.860} \\\hline
		DUT & 0.736 & 0.803 & 0.816 & \textbf{0.832} \\\hline
		Pascal-S & 0.702 & 0.815 & 0.742 & \textbf{0.854} \\\hline
	\end{tabular}
\end{table}

\begin{figure*}[htp]
    \centering
    \includegraphics[width=\textwidth]{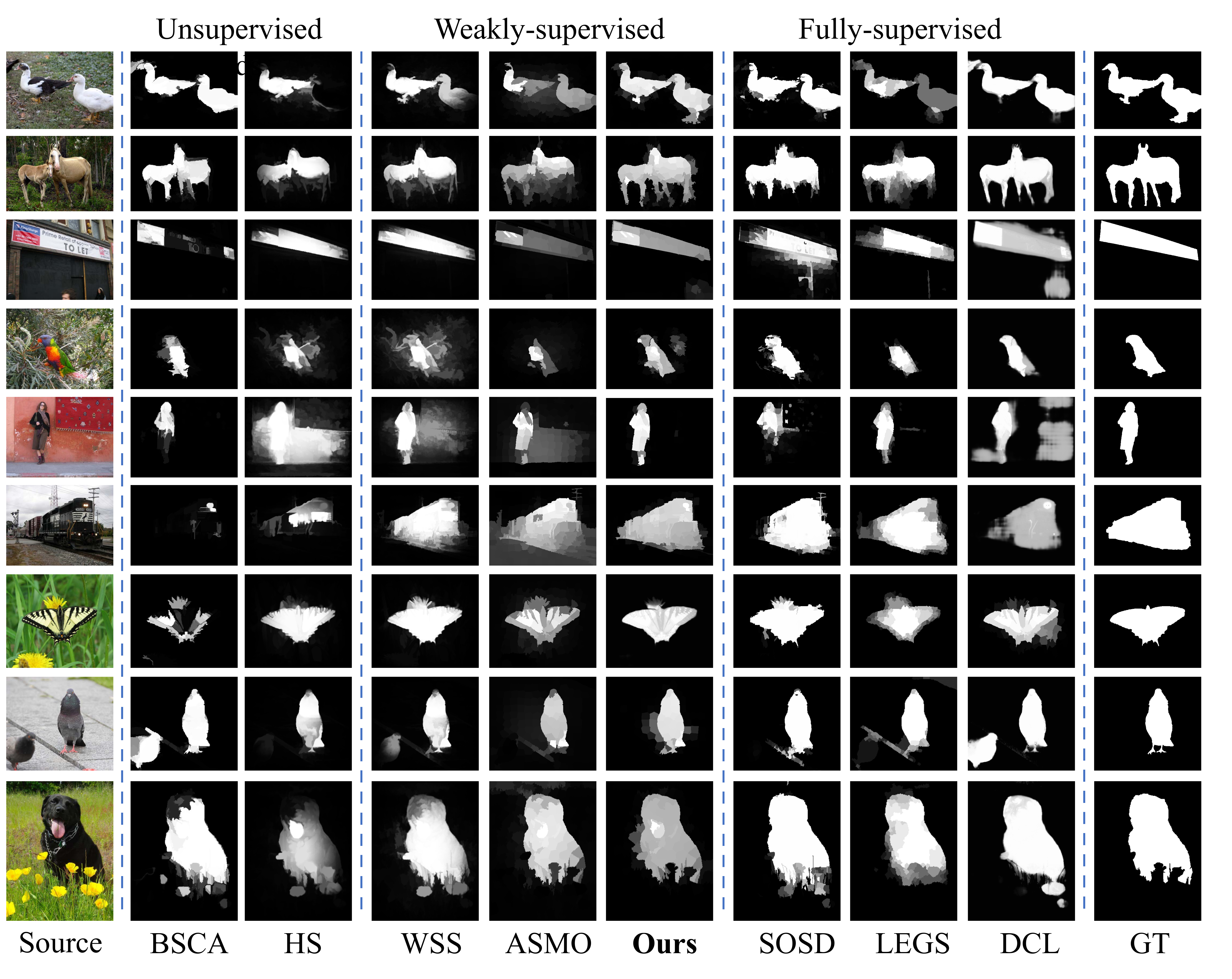}
    \caption{Qualitative comparison with other saliency detection methods. Unsupervised methods, weakly-supervised methods and fully-supervised methods are placed from left to right. Among those weakly-supervised methods, our proposed method produces saliency maps closest to the ground-truth masks.}
    \label{fig:comparison}
\end{figure*}

The PR curves on the ECSSD, HKU-IS and DUT-OMRON datasets are presented in Figure~\ref{fig:pr}. Our method consistently outperforms other unsupervised methods and weakly-supervised methods. It is also better than some fully-supervised methods like SOSD \cite{he2017delving}, LEGS \cite{wang2017learning} and DCL\cite{li2016deep}, except on the ECSSD dataset where ours is very close to the result of DCL. We also evaluate those methods on \textit{S-measure}. 
The results are shown in Table~\ref{tab:s}. It reveals that our method generates saliency maps of higher structural similarity compared with the ground-truth masks. The qualitative result is shown in Figure~\ref{fig:comparison}. The first two rows show that our method provides clear separation between multiple objects. The next four rows present that our results maintain the complete appearance of the salient objects. Moreover, we generate saliency maps with clear boundaries, as shown in the last three rows.

\begin{figure*}
    \centering
    \includegraphics[width=\textwidth]{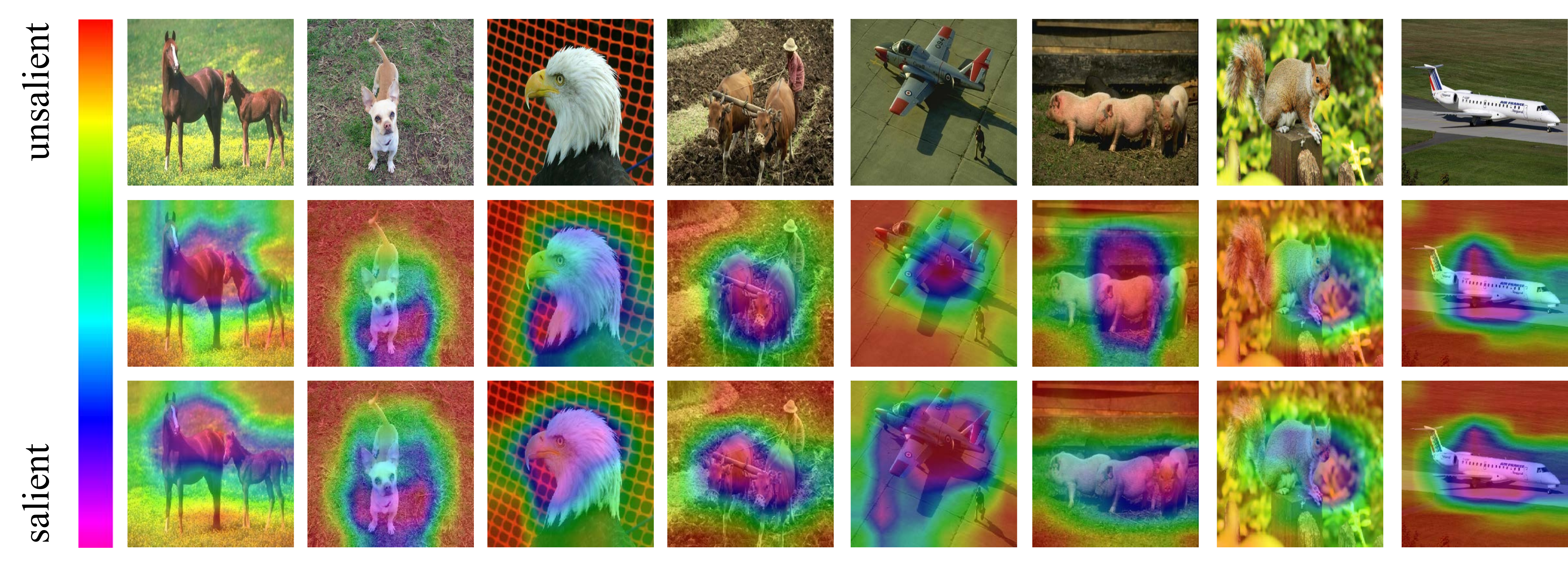}
    \caption{Comparison between different iterations. From top to bottom, they are original images, saliency maps generated with Grad-CAM, and saliency maps after 50 iterations. Saliency maps after 50 iterations cover larger activation area belonging to the salient objects. The color code on the left represents the degree of saliency.}
    \label{fig:map}
\end{figure*}

The superior performance of our proposed method confirms that object subitizing generalizes better to the saliency detection task than image-level supervision. We also conduct extensive experiments to validate the performance of each component in our method.

  

\section{Ablation Study}

In this section, we discuss the advantage of subitizing supervisions over image-level supervisions. We also evaluate the effectiveness of the Saliency Updating Module and the refinement process.

\subsection{The Advantage of Subitizing Supervisions}
\begin{figure}
    \centering
    \includegraphics[width=0.5\textwidth]{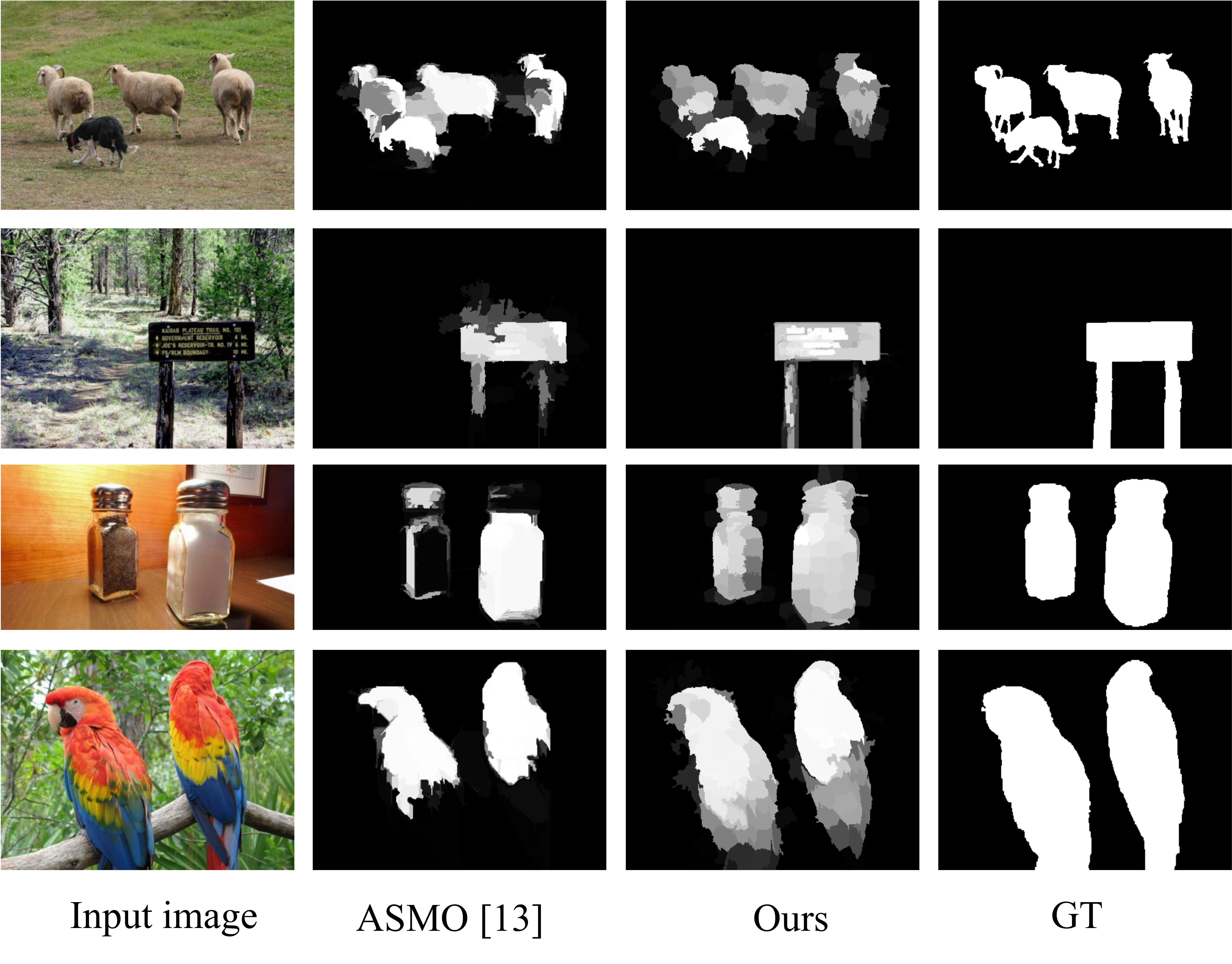}
    \caption{Comparison between our method and ASMO \cite{li2018weakly}, a weakly-supervised method using image-tag supervision. The results reveal the superiority of subitizing supervision.}
    \label{fig:sub}
\end{figure}

In this subsection, we aim to reveal the advantage of subitizing supervisions over image-tag supervisions. The generated saliency maps from our method are compared against those from ASMO \cite{li2018weakly} with image-label supervisions, as presented in Figure~\ref{fig:sub}. The first two rows reveal that the subitizing supervision helps recognize the border between multiple objects. The last two rows indicate that our method captures the whole regions of the salient objects, while those methods supervised with image tags leave out some parts of the salient objects. In addition, we train our network with the image-tag supervisions. The results are also process with the refinement module. The performance with different training data is presented in Figure \ref{tab:train_with_image}. 
The subitizing-supervised framework performs better than the image-tag-supervised framework, which reveals the advantage of subitizing supervision.

\begin{table}
    \centering
    \caption{The performance of our framework trained with image tags and subitizing, respectively. The better ones are marked in \textbf{bold}.}
    \label{tab:train_with_image}
    \begin{tabular}{|c|c|c|c|c|}
    \hline
        Dataset & Metric & w/ image-tag & w/ subitizing \\\hline
        \multirow{2}{*}{ECSSD} & $F_{\beta}\uparrow$  & 0.825 & \textbf{0.858}  \\
        \cline{2-4}
        ~ & MAE$\downarrow$  & 0.110 & \textbf{0.108}  \\\hline
        \multirow{2}{*}{DUT-OMRON} & $F_{\beta}\uparrow$  & 0.745 & \textbf{0.778}  \\
        \cline{2-4}
        ~ & MAE$\downarrow$  & 0.103 & \textbf{0.083}   \\\hline
    \end{tabular}
\end{table}

\subsection{The Effect of Saliency Updating Module}

In this subsection, we aim to evaluate the effect of the Saliency Updating Module. As shown in Figure~\ref{fig:map}, the results after updating are more complete in appearance, while those without any updating only focus on a limited but notable region due to the property of neural networks. With the SUM module, the network captures more parts within the semantic affinity. In addition, on the DUT-OMRON dataset, the $F_\beta$ and MAE measures of saliency predictions after 10 and 50 iterations with our SUM module are 0.638/0.252 and 0.704/0.139, respectively. It reveals that the SUM module helps boost the performance of saliency detection. 



\subsection{The Effect of Refinement Process}

In order to obtain promising results, we adopt the refinement process to optimize the boundaries of saliency maps. 
As CRF is the most popular technique to refine segmentation results, we apply CRF on coarse maps and evaluate the outputs as well. 
The results on the ECSSD, MSRA-B and Pascal-S datasets are presented in Table~\ref{tab:ablation}. The refinement process contributes a lot to the recognition of salient objects. It reveals that our refinement process achieves better optimization results than CRF. 
In addition, to evaluate the effectiveness of the refinement module on other methods, the refinement process is conducted on unsupervised results from BSCA \cite{qin2015saliency}. As shown in Table \ref{tab:unsuper}, the refinement module helps improve the performance, but the processed results are still worse than our results.

\begin{table}
    \centering
    \caption{The performance with/without the refinement process. The saliency results with CRF is also presented. The best performance is marked in \textbf{bold}.}
    \label{tab:ablation}
    \begin{tabular}{|c|c|c|c|c|c|}
    \hline
        Dataset & Metric & w/o post-pro. & w/ CRF & w/ ref. \\\hline
        \multirow{2}{*}{ECSSD} & $F_{\beta}\uparrow$  & 0.707 & 0.721 & \textbf{0.858} \\
        \cline{2-5}
        ~ & MAE$\downarrow$  & 0.197 & 0.185 & \textbf{0.108} \\\hline
        \multirow{2}{*}{MSRA-B} & $F_{\beta}\uparrow$  & 0.731 & 0.782 & \textbf{0.897} \\
        \cline{2-5}
        ~ & MAE$\downarrow$  & 0.167 & 0.152 & \textbf{0.082}  \\\hline
        \multirow{2}{*}{Pascal-S} & $F_{\beta}\uparrow$  & 0.644 & 0.680 & \textbf{0.803} \\
        \cline{2-5}
        ~ & MAE$\downarrow$  & 0.206 & 0.191 & \textbf{0.131}  \\\hline
    \end{tabular}
\end{table}

\begin{table}
    \small
    \centering
    \caption{The performance of unsupervised results (BSCA \cite{qin2015saliency}) processed by our refinement module.}
    \label{tab:unsuper}
    \begin{tabular}{|c|c|c|c|c|c|}
    \hline
        Dataset & Metric & BSCA\cite{qin2015saliency} & BSCA w/ ref. & ours w/ ref. \\\hline
        \multirow{2}{*}{ECSSD} & $F_{\beta}\uparrow$ & 0.705 & 0.756 & \textbf{0.858}  \\
        \cline{2-5}
        ~ & MAE$\downarrow$ & 0.183 & 0.140 & \textbf{0.108} \\\hline
        \multirow{2}{*}{DUT-O} & $F_{\beta}\uparrow$ & 0.500 & 0.618 & \textbf{0.778} \\
        \cline{2-5}
        ~ & MAE$\downarrow$ & 0.196 & 0.134 & \textbf{0.083}  \\\hline
    \end{tabular}
\end{table}


\section{Conclusion}
In this paper, we propose a novel method for the salient object detection task with the subitizing supervision. We design a model with the Saliency Subitizing Module and the Saliency Updating Module, which generates the initial masks using subitizing information and iteratively refines the generated saliency masks, respectively.
Without any seeds from unsupervised methods, our method outperforms other weakly-supervised methods and even performs comparable to some fully-supervised methods.

\appendices


\section*{Acknowledgment}

We thank for the support from National Natural Science Foundation of China(61972157, 61902129), Shanghai Pujiang Talent Program (19PJ1403100), Economy and Information Commission of Shanghai  (XX-RGZN-01-19-6348), National Key Research and Development Program of China (No. 2019YFC1521104), Science and Technology Commission of Shanghai Municipality Program (No. 18D1205903). Xin Tan is also supported by the Postgraduate Studentship (Mainland Schemes) from City University of Hong Kong.

\ifCLASSOPTIONcaptionsoff
  \newpage
\fi



\bibliographystyle{IEEEtran}
\end{document}